\documentclass[a4paper,twoside]{article}

\usepackage{epsfig}
\usepackage{subcaption}
\usepackage{calc}
\usepackage{amssymb}
\usepackage{amstext}
\usepackage{amsmath}
\usepackage{amsthm}
\usepackage{multicol}
\usepackage{pslatex}
\usepackage{apalike}
\usepackage{algorithm2e}
\usepackage[bottom]{footmisc}

\usepackage{graphicx}
\usepackage{fancyhdr}

\usepackage{hyperref}
\usepackage[capitalize]{cleveref}
\usepackage{csquotes}
\usepackage{multirow}
\usepackage{siunitx}
\usepackage{booktabs}
\usepackage{nicefrac}
\usepackage{xcolor}

\hypersetup{%
    pdftoolbar=true,        
    pdfmenubar=true,        
    pdffitwindow=false,     
    pdfstartview={FitH},    
    pdftitle={Efficient Neural Network Training via Subset Pretraining},    
    pdfauthor={Jan Spörer, Bernhard Bermeitinger, Tomas Hrycej, Niklas Limacher, Siegfried Handschuh},     
    pdfkeywords={Deep Neural Network} {Convolutional Network} {Computer Vision} {Efficient Training} {Resource Optimization} {Training Strategies} {Determination Ratio} {Stochastic Approximation Theory}, 
    pdfnewwindow=true,      
    breaklinks=true,
}

\usepackage{SCITEPRESS}

\newcommand{\mnist}{\mbox{MNIST}}
\newcommand{\cifar}{\mbox{CIFAR-10}}
\newcommand{\cifarb}{\mbox{CIFAR-100}}

\begin{document}

\title{Efficient Neural Network Training via Subset Pretraining}

\author{%
	\authorname{%
		Jan Spörer\sup{\dagger}\sup{1},
		Bernhard Bermeitinger\sup{\dagger}\sup{2},
		Tomas Hrycej\sup{\dagger}\sup{1},
		Niklas Limacher\sup{\ddagger}\sup{1},
		and Siegfried Handschuh\sup{\dagger}\sup{1}  
	}
	\affiliation{%
		\sup{1}Institute of Computer Science, University of St.Gallen (HSG), St.Gallen, Switzerland
	}
	\affiliation{%
		\sup{2}Institute of Computer Science in Vorarlberg, University of St.Gallen (HSG), Dornbirn, Austria
	}
	\email{%
        \sup{\dagger}firstname.lastname@unisg.ch,
        \sup{\ddagger}firstname.lastname@student.unisg.ch
    }
}


\abstract{%
	In training neural networks, it is common practice to use partial gradients computed over batches, mostly very small subsets of the training set.
	This approach is motivated by the argument that such a partial gradient is close to the true one, with precision growing only with the square root of the batch size.
	A theoretical justification is with the help of stochastic approximation theory.
	However, the conditions for the validity of this theory are not satisfied in the usual learning rate schedules.
	Batch processing is also difficult to combine with efficient second-order optimization methods.
	This proposal is based on another hypothesis: the loss minimum of the training set can be expected to be well-approximated by the minima of its subsets.
	Such subset minima can be computed in a fraction of the time necessary for optimizing over the whole training set.
	This hypothesis has been tested with the help of the \mnist, \cifar{}, and \cifarb{} image classification benchmarks, optionally extended by training data augmentation.
	The experiments have confirmed that results equivalent to conventional training can be reached.
    In summary, even small subsets are representative if the overdetermination ratio for the given model parameter set sufficiently exceeds unity.
	The computing expense can be reduced to a tenth or less.
}

\onecolumn \maketitle \normalsize \setcounter{footnote}{0} \vfill

\pagestyle{fancy}
\fancyhf{} 
\fancyfoot[L]{To appear in KDIR 2024.}
\renewcommand{\headrulewidth}{0pt} 
\renewcommand{\footrulewidth}{0pt} 

\section{\uppercase{Introduction}}\label{sec:introduction}
Neural networks as forecasting models learn by fitting the model forecast to the desired reference output (e.g., reference class annotations) given in the data collection called the training set.
The fitting algorithm changes model parameters in the loss function's descent direction, measuring its forecast deviation.
This descent direction is determined using the loss function gradient.

The rapidly growing size of neural networks (such as those used for image or language processing) motivates striving for a maximum computing economy.
One widespread approach is determining the loss function gradient from subsets of the training set, called batches (or mini-batches).
Different batches are alternately used to cover the whole training set during the training.

There are some arguments supporting this procedure.
\cite[Section 8.1.3]{goodfellow_deep_2016} refers to the statistical fact that random standard deviation decreases with the square root of the number of samples.
Consequently, the gradient elements computed from a fraction of $ \nicefrac{1}{K} $ training samples (with a given positive integer $K$) have a standard deviation equal to the factor $\sqrt{K}$ multiple of those computed over the whole training set, which seems to be a good deal.

Another frequent justification is with the help of stochastic approximation theory.
The stochastic approximation principle applies when drawing training samples from a stationary population generated by a fixed (unknown) model.
\cite{robbins_stochastic_1951} discovered this principle in the context of finding the root (i.e., the function argument for which the function is zero) of a function $ g(x) $ that cannot be directly observed.
What can be observed are randomly fluctuating values $ h(x) $ whose mean value is equal to the value of the unobservable function, that is,
\begin{equation}\label{eq:robin_munro_ehx}
	E \left[ h(x) \right] = g(x)
\end{equation}

The task is to fit an input/output mapping to data by gradient descent.
For the parameter vector $w$ of this mapping, the mean of the gradient $h(w)$ with respect to the loss function computed for a single training sample is expected to be equal to the gradient $g(w)$ over the whole data population.
The local minimum of the loss function is where the gradient $g(w)$ (i.e., the mean value of $h(w)$) is zero.
\cite{robbins_stochastic_1951} have proven that, under certain conditions, the root is found with probability one (but without a concrete time upper bound).

However, this approach has some shortcomings.
For different batches, the gradient points in different directions.
So, the descent for one batch can be an ascent for another.
To cope with this, \cite{robbins_stochastic_1951} formulated the convergence conditions.
They require that if the update rule for the parameter vector is
\begin{equation}\label{eq:robin_munro_update}
	w_{t+1} = w_t - c_t h(w_t)
\end{equation}
which corresponds to the usual gradient descent with step size $c_t$, the step size sequence $c_t$ has to satisfy the following conditions:
\begin{equation}\label{eq:robin_munro_ct}
	\sum_{t=1}^\infty c_t = \infty
\end{equation}
and
\begin{equation}\label{eq:robin_munro_ct2}
	\sum_{t=1}^\infty c_t^2 < \infty
\end{equation}

Condition~\cref{eq:robin_munro_ct} is necessary for the step not to vanish prematurely before reaching the optimum with sufficient precision.
Condition~\cref{eq:robin_munro_ct2} provides for decreasing step size.
With a constant step size, the solution would infinitely fluctuate around the optimum.
This is because, in the context of error minimization, its random instance $ h(w) $ will not diminish for individual samples, although the gradient $ g(w) = E[ h(x) ] $ will gradually vanish as it approaches the minimum.
Finally, the gradients of individual samples will not vanish even if their mean over the training set is zero.
At the minimum, $ g(w) = 0 $ will result from a balance between individual nonzero vectors $ h(w) $ pointing to various directions.

\section{\uppercase{Shortcomings of the batch oriented approach}}\label{sec:shortcomings}
The concept of gradient determination using a subset of the training set is mostly satisfactory.
However, several deficiencies from theoretical viewpoints suggest an enhancement potential.

\subsection{\uppercase{Violating the conditions of the stochastic approximation}}\label{sec:violating_stoch_approx}
The conditions~\cref{eq:robin_munro_ct} and~\cref{eq:robin_munro_ct2} for convergence of the stochastic approximation procedure to a global (or at least local) minimum result from the stochastic approximation theory.
Unfortunately, they are almost always neglected in the neural network training practice.
This may lead to a bad convergence (or even divergence).
The common \emph{Stochastic Gradient Descent}~(SGD) with a fixed learning step size violates the stochastic approximation principles.
However, even popular sophisticated algorithms do not satisfy the conditions.
The widespread (and successful) \emph{Adam}~\cite{kingma_adam_2015} optimizer uses a weight consisting of the quotient of the exponential moving average derivative and the exponential moving average of the square of the derivative
\begin{equation}\label{eq:adam}
	\begin{split}
		w_{t+1, i} & =
		w_{t,i}
		- \frac{c m_{t, i}}
		{\sqrt{d_{t, i}}}
		\frac{\partial E \left( w_{t, i} \right)}
		{\partial w_{t, i}}   \\
		m_{t,i}    & =
		\beta_1 d_{t-1, i}
		+ \left( 1 - \beta_1 \right)
		\frac{\partial E \left( w_{t-1, i} \right)}
		{\partial w_{t-1, i}} \\
		d_{t, i}   & =
		\beta_2 d_{t-1, i}
		+ \left( 1 - \beta_2 \right)
		{\left( \frac{\partial E \left( w_{t-1, i} \right)}
			{\partial w_{t-1, i}}
			\right)}^2
	\end{split}
\end{equation}
with metaparameters $c$, $\beta_1$, and $\beta_2$, network weights $w_{t,i}$, and the loss function $E$.
$\beta_1$ is the decay factor of the exponential mean of the error derivative, $\beta_2$ is the decay factor of the square of the error derivative, and $c$ is the step length scaling parameter.
Their values have been set to the framework's sensible defaults $ c = 0.001 $, $ \beta_1 = 0.9 $, and $ \beta_2 = 0.999 $ in the following computing experiments.

Normalizing the gradient components by the moving average of their square via $\sqrt{d_{t, i}}$ is the opposite of the decreasing step size required in the stochastic approximation theory.
If the gradient becomes small (as expected at the proximity of the minimum), the normalization increases them.
This may or may not be traded off by the average gradient $m_{t,i}$.

\subsection{\uppercase{\enquote{Good} Approximation of Gradient is not Sufficient}}\label{approx_grad}

The quoted argument of~\cite[Section 8.1.3]{goodfellow_deep_2016} that the standard deviations of gradient components are decreasing with the square root $\sqrt{K}$ of number of samples used while the computing expense is increasing with $K$ is, of course, accurate for independent samples drawn from a population.

However, this relationship is only valid if the whole statistical population (from which the training set is also drawn) is considered.
This does not account for the nature of numerical optimization algorithms.
The more sophisticated among them follow the descent direction.
The gradient's statistical deviation is relatively small with respect to the statistical population, but this does not guarantee that a so-determined estimated descent direction is not, in fact, an ascent direction.
The difference between descent and ascent may easily be within the gradient estimation error --- the batch-based gradient is always a sample estimate, with standard deviation depending on the unknown variance of the individual derivatives within the training set.
By contrast, optimizing over the training set itself, the training set gradient is computed deterministically, with zero deviation.
Then, the descent direction is certain to lead to a decrease in loss.

The explicit task of the optimization algorithm is to minimize the loss over the training set.
If the goal of optimizing over the whole (explicitly unknown) population is adopted, the appropriate means would be biased estimates that can have lower errors over the population, such as ridge regression for linear problems~\cite{van_wieringen_lecture_2023}.
The biased estimate theory provides substantial results concerning this goal but also shows that it is difficult to reach because of unknown regularization parameters, which can only be determined with computationally expensive experiments using validation data sets.

Even if the loss is extended with regularization terms to enhance the model's performance on the whole population (represented by a validation set), the optimized regularized fit is reached at the minimum of the extended loss function once more over \emph{the given training set}.
Thus, as mentioned above, it is incorrect from the optimization algorithm's viewpoint to compare the precision of the training set gradient with that of the batches, which are subsamples drawn from the training set.
The former is precise, while the latter are approximations.

The related additional argument frequently cited is that what is genuinely sought is the minimum for the population and not for the training set.
However, this argument is somewhat misleading.
There is no method for finding the true, exact minimum for the population only based on a subsample such as the training set --- the training set is the best and only information available.
Also, the loss function values used in the algorithm to decide whether to accept or reject a solution are values for the given training set.
Examples in~\cite{hrycej_mathematical_2023} show that no law guarantees computing time savings through incremental learning for the same performance.

\subsection{\uppercase{Convexity around the minimum is not exploited}}\label{sec:convexity}

Another problem is that in a specific environment of the local minimum, every smooth function is convex --- this directly results from the minimum definition.
Then, the location of the minimum is not determined solely by the gradient; the Hessian matrix also captures the second derivatives.
Although using an explicit estimate of the Hessian is infeasible for large problems with millions to billions of parameters, there are second-order algorithms that exploit the curvature information implicitly.
One of them is the well-known conjugate gradient algorithm~\cite{hestenes_methods_1952,fletcher_function_1964}, thoroughly described in~\cite{press_numerical_1992}, which requires only the storage of an additional vector with a dimension equal to the length of the plain gradient.
However, batch sampling substantially distorts the second-order information more than the gradient~\cite{goodfellow_deep_2016}.
This leads to a considerable loss of efficiency and convergence guarantee of second-order algorithms, which is why they are scarcely used in the neural network community, possibly sacrificing the benefits of their computing efficiency.

Second-order algorithms cannot be used with the batch scheme for another reason.
They are usually designed for continuous descent of loss values.
Reaching a specific loss value with one batch cannot guarantee that this value will not become worse with another batch.
This violates some assumptions for which the second-order algorithms have been developed.
Mediocre computing results with these algorithms in the batch scheme seem to confirm this hypothesis.

\section{\uppercase{Substituting the training set by a subset}}\label{sec:substituting_subset}

To summarize the arguments in favor of batch-oriented training, the batch-based procedure is justified by the assumption that the gradients for individual batches are roughly consistent with the gradient over the whole training set (epoch).
So, a batch-based improvement is frequently enough (but not always, depending on the choice of the metaparameters) also an improvement for the epoch.
This is also consistent with the computing experience record.
On the other hand, one implicitly insists on optimizing over the whole training set to find an optimum, as one batch is not expected to represent the training set fully.

Batch-oriented gradient optimization hypothesizes that the batch-loss gradient approximates the training set gradient and the statistic population gradient well enough.

By contrast, the hypothesis followed here is related but essentially different.
It is assumed that \emph{the optimum of the loss subset is close to the optimum of the training set}.

\begin{figure}
	\centering
	\includegraphics[width=\columnwidth]{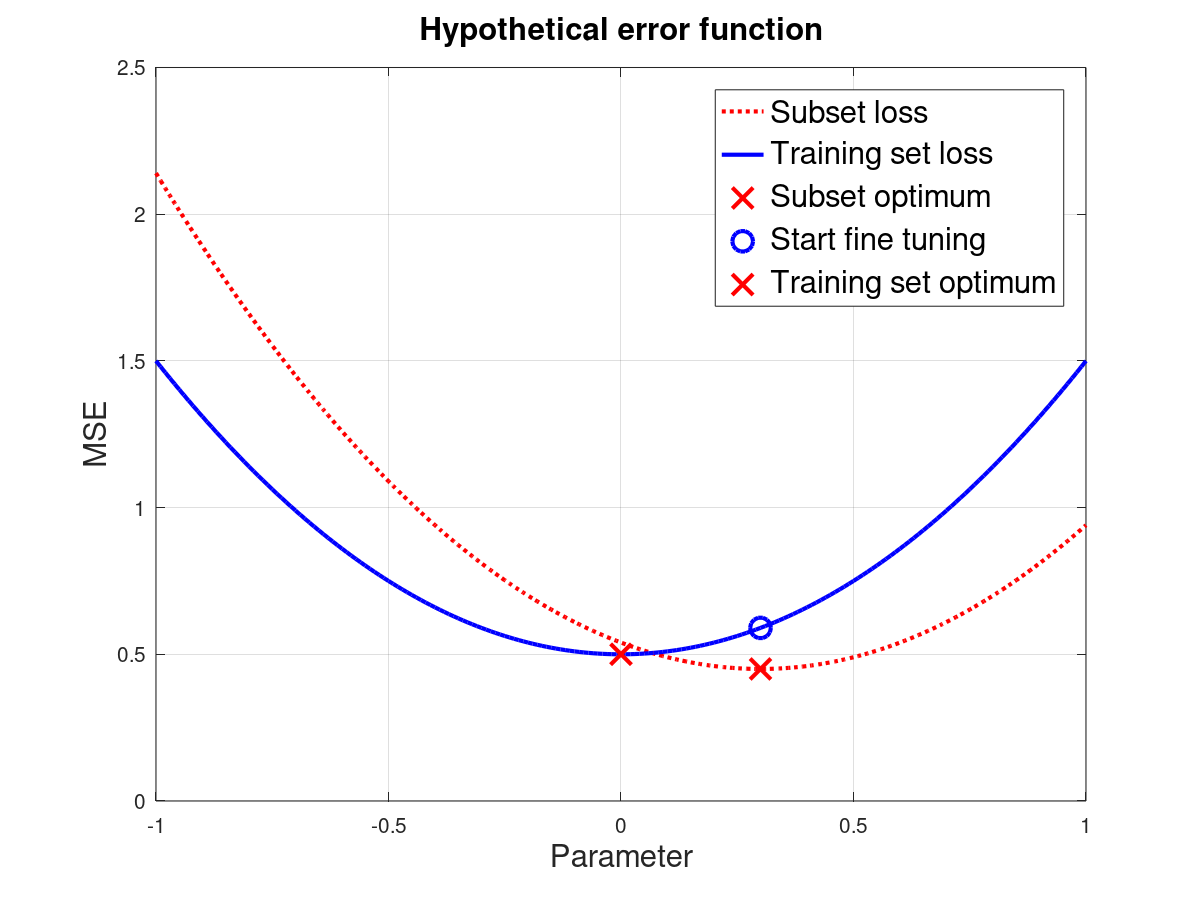}
	\caption{Optima for a subset and the whole training set.}\label{fig:subset_training}
\end{figure}

Even if the minimum approximation is imperfect, it can be expected to be a very good initial point for fine-tuning the whole training set so that a few iterations may suffice to reach the true minimum.
This principle is illustrated in~\cref{fig:subset_training}.
The subset loss function (red, dotted) is not identical to the training set loss function (blue, solid).
Reaching the minimum of the subset loss function (red cross) delivers an initial point for fine-tuning on the training set (blue circle).
This initial point is close to the training set loss minimum (blue cross) and is very probably within a convex region around the minimum.
This motivates using fast second-order optimization methods such as the conjugate gradient method~\cite{press_numerical_1992}.

Another benefit of such a procedure is that for a fixed subset, both the gradient and the intermediary loss values are exact.
This further justifies the use of second-order optimization methods.

Of course, the question is how large the subset has to be for the approximation to be sufficiently good.
As previously noted, a smooth function is always locally convex around a minimum.
If the approximate minimum over the training subset is in this environment, conditions for efficient minimization with the help of second-order algorithms are satisfied.
Then, a fast convergence to the minimum over the whole training set can be expected.

Consequently, it would be desirable for the minimum of the subset loss to be within the convex region of the training set loss.

\section{\uppercase{Setup of computing experiments}}\label{sec:computing_experiments}

The following computing experiments investigate the support of these hypotheses. The experimental method substitutes the training set with representative subsamples of various sizes.
Subsequently, a short fine-tuning on the whole training set has been performed to finalize the optimum solution.
The model is trained on the subsamples for exactly \num{1 000} epochs, while the fine-tuning on the whole training set is limited to \num{100} epochs.

\subsection{\uppercase{Benchmark Datasets Used}}

The benchmarks for the evaluation have been chosen from the domain of computer vision.
They are medium-sized problems that can be run for a sufficient number of experiments.
This would not be possible with large models such as those used in language modeling.

For the experiments, three well-known image classification datasets \mnist{}~\cite{lecun_gradient-based_1998}, \cifar{}, and \cifarb{}~\cite{krizhevsky_learning_2009} were used.
\mnist{} contains grayscale images of handwritten digits (0--9) while \cifar{} contains color images of exclusively ten different mundane objects like \enquote{horse}, \enquote{ship}, or \enquote{dog}.
\cifarb{} contains the same images; however, they are classified into \num{100} fine-grained classes.
They contain \num{60 000} (\mnist{}) and \num{50 000} (\cifar{} and \cifarb{}) training examples.
Their respective preconfigured test split of each \num{10 000} examples are used as validation sets.
While both \cifar{} and \cifarb{} are evenly distributed among the classes, \mnist{} is roughly evenly distributed.
We opted to proceed without mitigating the slight class imbalance.

\subsection{\uppercase{The Model Architecture}}

The model is a convolutional network inspired by the \emph{VGG} architecture~\cite{simonyan_very_2015}.
It uses three consecutive convolutional layers with the same kernel size of $3 \times 3$, 32/64/64 filters, and the ReLU activation function.
Each is followed by a maximum pooling layer of size $2 \times 2$.
The last feature map is flattened, and after a classification block with one layer of 64 units and the ReLU activation function, a linear dense layer classifies it into a softmax vector.
All trainable layers' parameters are initialized randomly from the Glorot uniform distribution~\cite{glorot_understanding_2010} with a unique seed per layer such that all trained models throughout the experiments have an identical starting point.
The biases of each layer are initialized by zeros.
The number of parameters for the models differs only because \mnist{} has one input channel, while \cifar{} and \cifarb{} have three, and \cifarb{} has 100 class output units instead of 10.

\subsection{\uppercase{Preventing Underdetermination of Model Parameters}}

An important criterion is that the training set size is sufficient for this procedure.
The size of the training subsets (as related to the number of model parameters) must be large enough for the model not to be underdetermined.
This should be true for most of the subsets tested so that we can fairly compare subsets that are a relatively small fraction of the training set.
As a criterion for this, the overdetermination ratio of each benchmark candidate has been evaluated~\cite{hrycej_mathematical_2023}:
\begin{equation}\label{eq:q_coeff}
	Q = \frac{KM}{P}
\end{equation}
with $K$ being the number of training examples, $M$ being the output vector length (usually equal to the number of classes), and $P$ being the number of trainable model parameters.

This formula justifies itself by ensuring that the numerator $KM$ equals the number of constraints to be satisfied (the reference values for all training examples).
This product must be larger than the number of trainable parameters for the system to be sufficiently determined.
(Otherwise, there are infinite solutions, most of which do not generalize.)
This is equivalent to the requirement for the overdetermination ratio $Q$ to be larger than unity.
It is advisable that this is satisfied for the training set subsets considered, although subsequent fine-tuning on the whole training set can \enquote{repair} a moderate underdetermination.

The two datasets \mnist{} and \cifar{} have ten classes.
This makes the number of constraints $KM$ in~\cref{eq:q_coeff} too small for subsets with $b > 4$.
This is why these two datasets have been optionally expanded by image augmentation.
This procedure implies slight random rotations, shifts, and contrast variations.
So, the number of training examples has been increased tenfold by augmenting the training data.
With \cifarb{} containing \num{100} classes, this problem does not occur, and it was not augmented.

\subsection{\uppercase{Processing Steps}}
The processing steps for every given benchmark task and a tested algorithm have been the following:
\begin{itemize}
	\item
	      The number of subsets $b$ such that a subset is the fraction $ \nicefrac{1}{b} $ of the training set has been defined.
	      These numbers have been: $b \in B$ with $B = \left\{ 2, 4, 8, 16, 32, 64, 128 \right\}$.
	      With a training set size $K$, a subset contains $ \nicefrac{K}{b} $ samples.
				For example, a value of $b=2$ results in a subset with half of the samples from the original training set.
	\item
	      All $b$ subsets of size $ \nicefrac{K}{b} $ have been built to support the results statistically.
	      Each subset $B_{bi}, i = 1, \ldots, b$, consists of training samples with index $i$ selected so that the subsets partition the entire training set.
	      The number of experiments is excessive for larger values of $b$, so only five random subsets are selected.
	      All randomness is seeded such that each experiment receives the same subset.
	\item
	      For every $ b \in B $ and every $ i = 1, \ldots, b $, the subset loss $ E_{bi} $ has been minimized using the selected training algorithm.
	      The number of epochs has been set to \num{1 000}.
	      Additionally, the losses for the whole training set $(E_{BTbi})$ and validation set ($E_{BVbi}$) have been evaluated.
	      Subsequently, a fine-tuning on the whole training set for \num{100} epochs has been performed, and the metrics for the training set ($E_{Tbi}$) and validation set $(E_{Vbi})$ have been evaluated.
	      In summary, the set of loss characteristics $E_{bi}$, $E_{BTbi}$, $E_{BVbi}$, $E_{Tbi}$, and $E_{Vbi}$ have represented the final results.
	\item
	      For comparison, the typical training on the original training set is given by choosing $b = 1$.
\end{itemize}

The conjugate gradient algorithm would be the favorite for optimizing the subset (because of its relatively small size) and fine-tuning (because of its expected convexity in the region containing the initial point delivered by the subset training).
Unfortunately, this algorithm is unavailable in deep learning frameworks like \emph{Keras}.
This is why the popular Adam algorithm has been used.
For reproducibility and removing additional hyperparameters, a fixed learning rate of \num{0.001} was employed for all training steps.

\section{\uppercase{Results}}\label{sec:results}
\subsection{\uppercase{Dataset \mnist{}}}\label{sec:results_mnist}
\begin{figure}
	\centering
	\includegraphics[height=5cm]{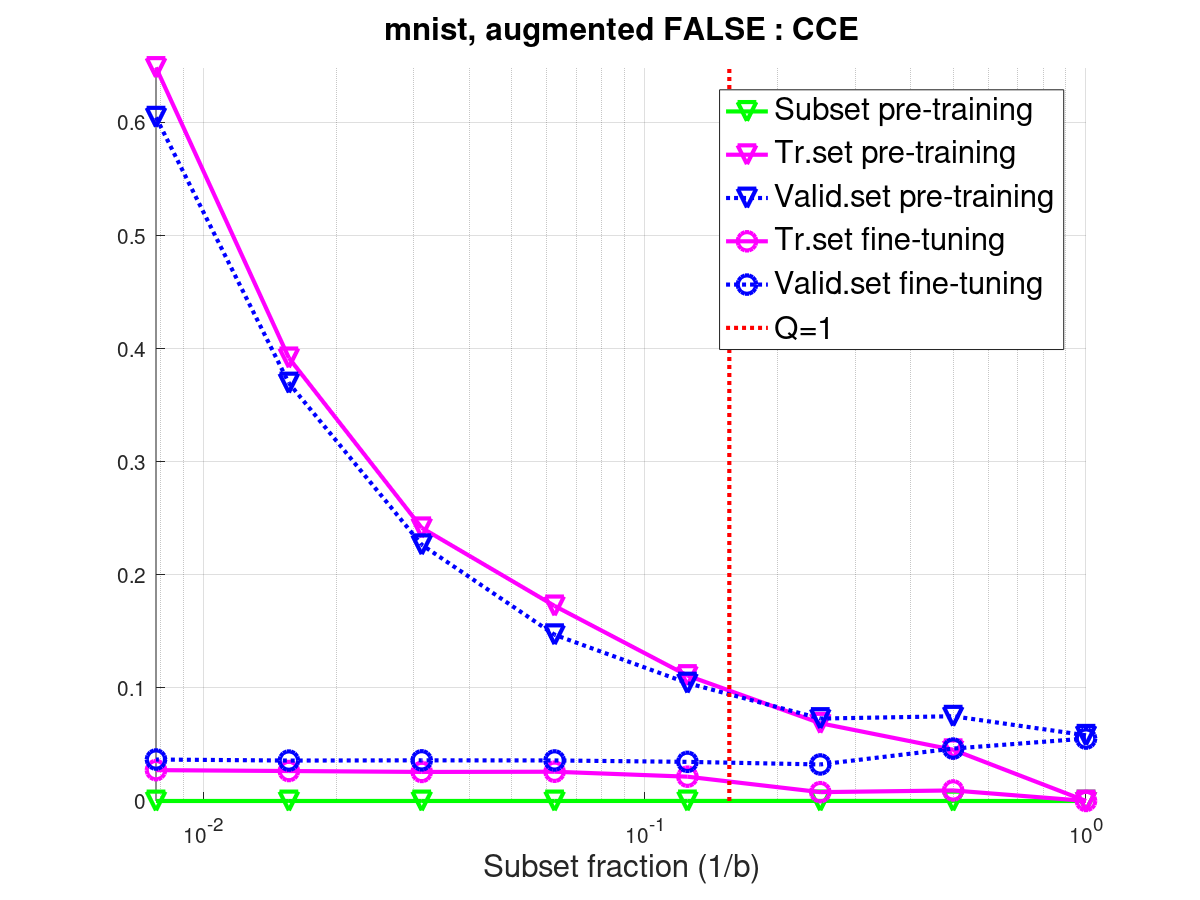}
	\caption{Dataset \mnist{} (not augmented), loss optima for a pre-trained subset and the whole training set in dependence from the subset size (as a fraction of the training set).}\label{fig:Loss_Mnist_naug}
\end{figure}

\begin{figure}
	\centering
	\includegraphics[height=5cm]{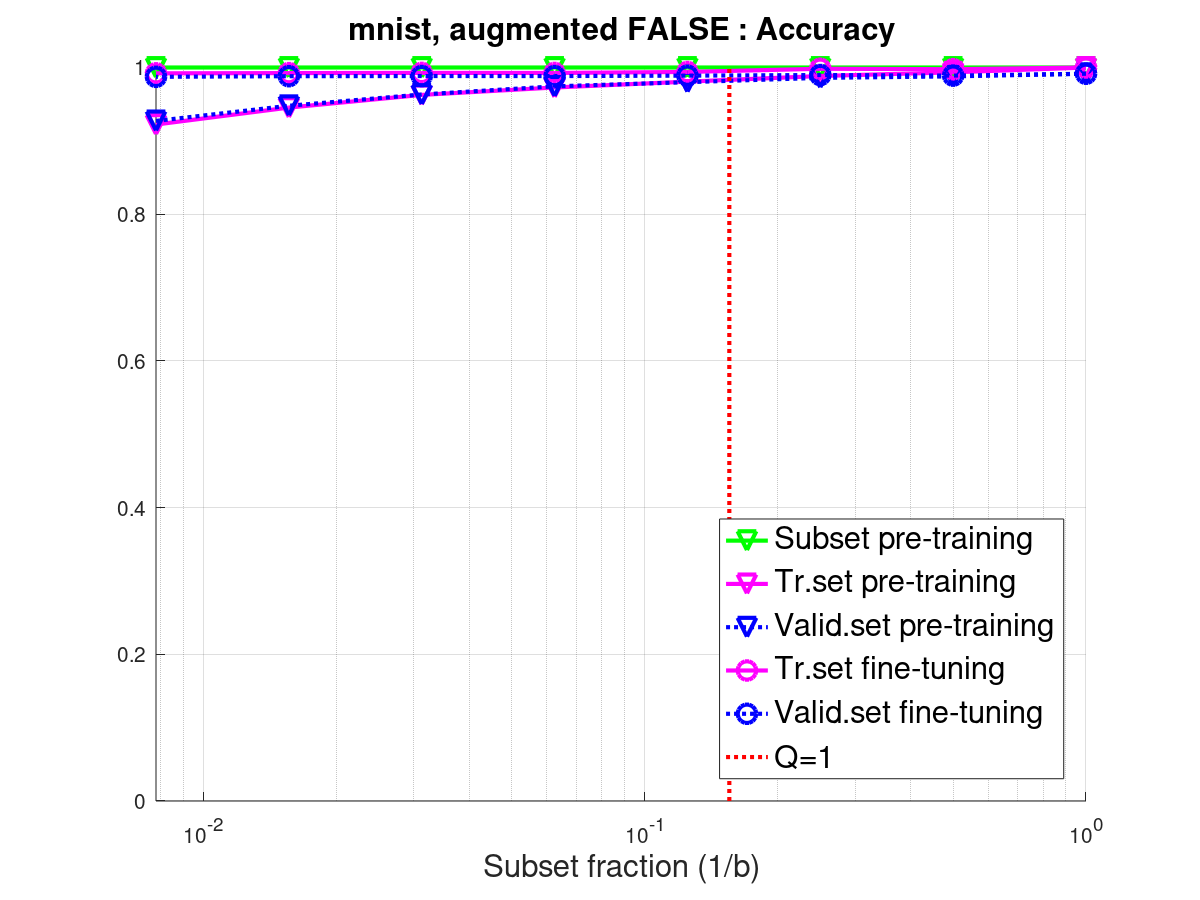}
	\caption{Dataset \mnist{} (not augmented), accuracy optima for a pre-trained subset and the whole training set in dependence from the subset size (as a fraction of the training set).}\label{fig:Accuracy_Mnist_naug}
\end{figure}

The results for the non-augmented \mnist{} dataset are depicted in~\cref{fig:Loss_Mnist_naug,fig:Accuracy_Mnist_naug}. 
The training has two phases:
\begin{enumerate}
	\item
	      the \emph{subset training} phase, in which only a fraction of the training set is used; and
	\item
	      the \emph{fine-tuning} phase, in which the optimized parameters from the subset training phase are fine-tuned by a (relatively short) training over the whole training set.
\end{enumerate}
On the $x$-axis, fractions of the complete training set are shown as used for the subset training.
The axis is logarithmic, so the variants are equally spaced.
These are $\nicefrac{1}{2}$, $\nicefrac{1}{4}$, $\nicefrac{1}{8}$, $\nicefrac{1}{16}$, $\nicefrac{1}{32}$, $\nicefrac{1}{64}$, and $\nicefrac{1}{128}$, as well as the baseline (fraction equal to unity).
This baseline corresponds to the conventional training procedure over the full training set.

The plotted magnitudes in~\cref{fig:Loss_Mnist_naug,fig:Accuracy_Mnist_naug} refer to
\begin{itemize}
	\item
	      the loss or accuracy reached for the given subset (\textit{Subset pre-training});
	\item
	      the loss or accuracy over the whole training set in the subset training optimum (\textit{Tr.set pre-training});
	\item
	      the loss or accuracy over the validation set in the subset training optimum (\textit{Valid.set pre-training});
	\item
	      the loss or accuracy over the whole training set attained through fine-tuning (\textit{Tr.set fine-tuning}); and
	\item
	      the loss or accuracy over the validation set in the fine-tuning optimum (\textit{Valid.set fine-tuning}).
\end{itemize}
All of them are average values over the individual runs with disjoint subsets.

The dotted vertical line marks the subset fraction with overdetermination ratio~\cref{eq:q_coeff} equal to unity.
To the left of this line, the subsets are underdetermined; to the right, they are overdetermined.

Both loss (\cref{fig:Loss_Mnist_naug}) and accuracy (\cref{fig:Accuracy_Mnist_naug}) suggest similar conjectures:
\begin{itemize}
	\item
	      The subset training with small subsets leads to poor training set and validation set losses.
	      This gap diminishes with the growing subset fraction.
	\item
	      Fine-tuning largely closes the gap between the training and validation sets.
	      The optimum value for the training set tends to be lower for large fractions (as they have an \enquote{advance} from the subset training, but this does not lead to a better validation set performance.
	      The baseline loss (the rightmost point) exhibits the highest validation set loss.
\end{itemize}
The overdetermination ratio delivers, together with the mentioned vertical lines in~\cref{fig:Loss_Mnist_naug} and~\cref{fig:Accuracy_Mnist_naug}, an additional finding: 
The gap between the performance on the subset and on the whole training set after the subset training is very large for $Q<1$ (the left side of the plot) and shrinks for $Q>1$ (the right side).

\begin{figure}
	\centering
	\includegraphics[height=5cm]{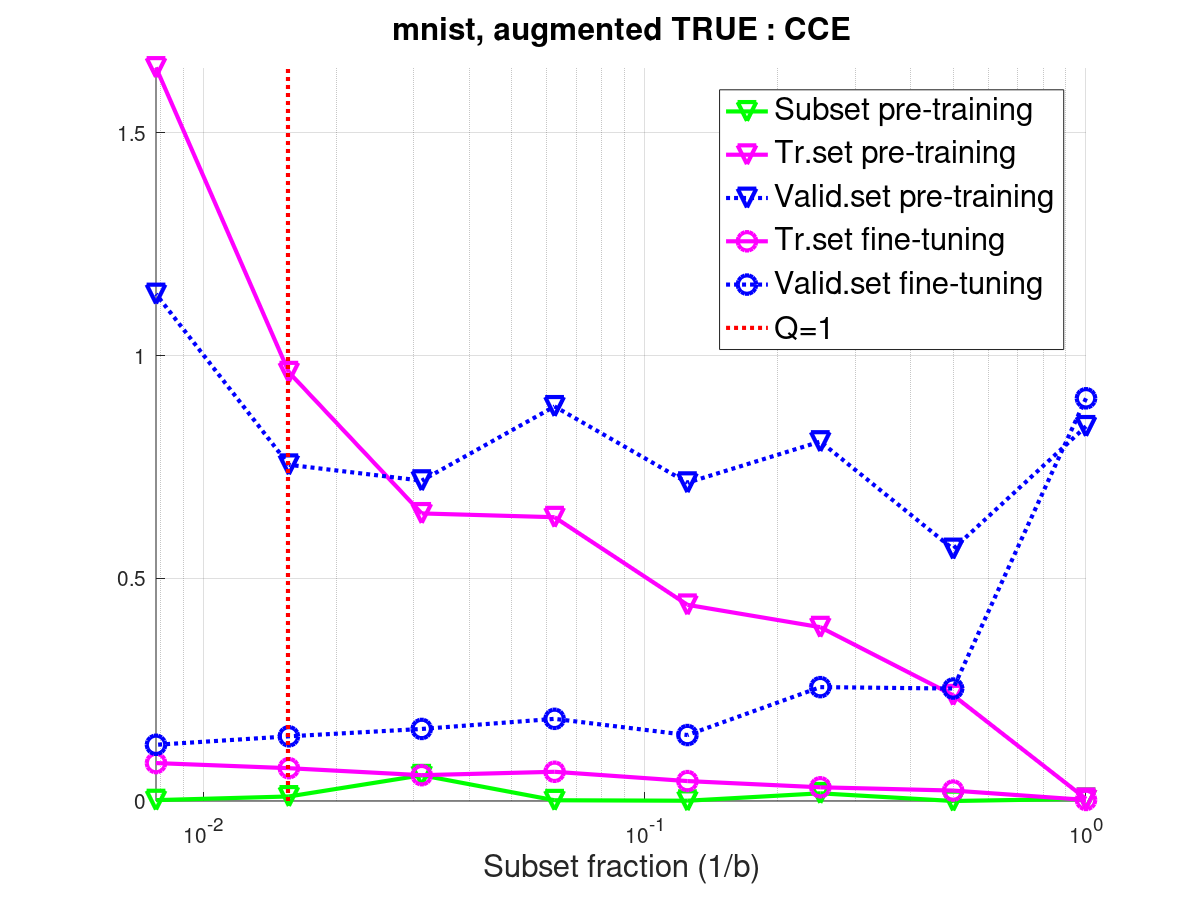}
	\caption{Dataset \mnist{} (augmented), loss optima for a pre-trained subset and the whole training set in dependence from the subset size (as a fraction of the training set).}\label{fig:Loss_Mnist_aug}
\end{figure}

\begin{figure}
	\centering
	\includegraphics[height=5cm]{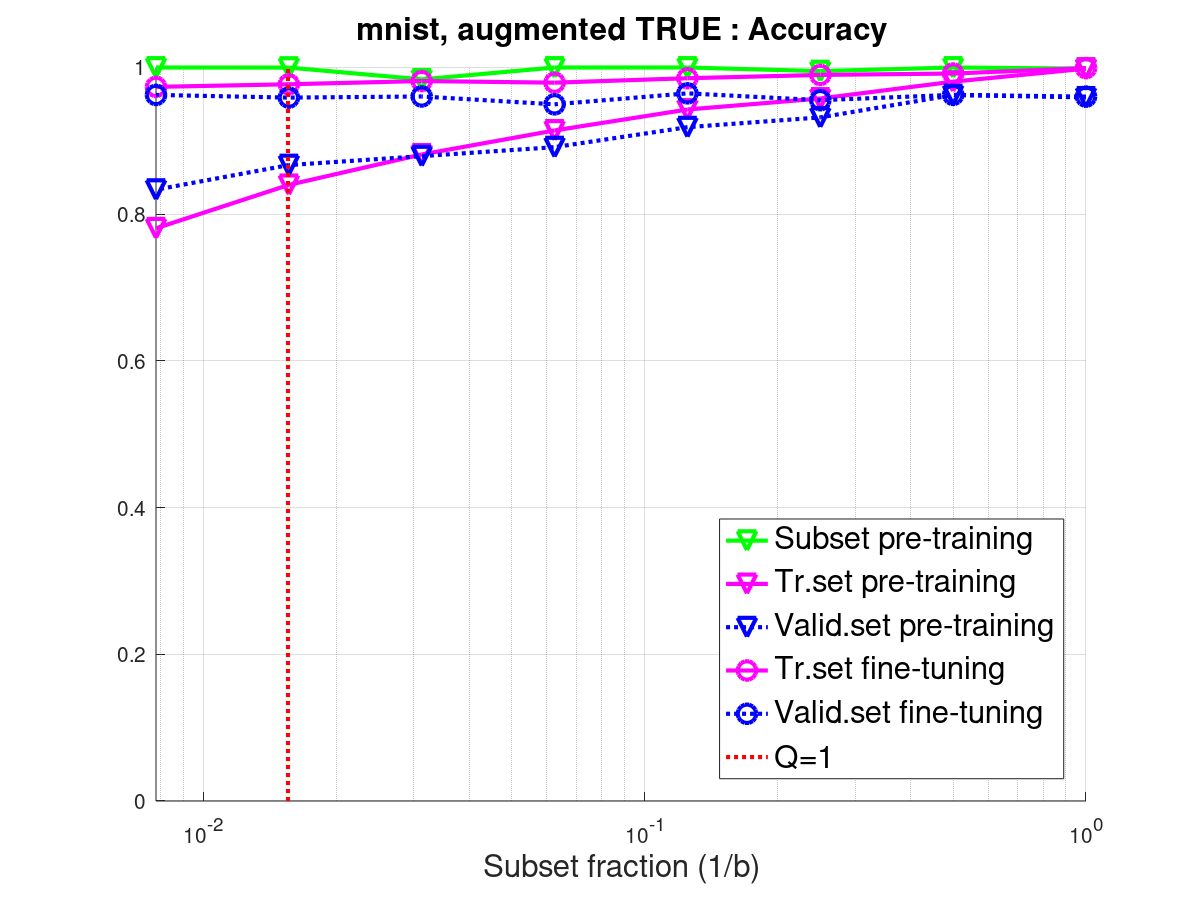}
	\caption{Dataset \mnist{} (augmented), accuracy optima for a pre-trained subset and the whole training set in dependence from the subset size (as a fraction of the training set).}\label{fig:Accuracy_Mnist_aug}
\end{figure}

The results for the augmented data are depicted in the plots~\cref{fig:Loss_Mnist_aug} and~\cref{fig:Accuracy_Mnist_aug}.
As the augmented data are more challenging to fit, their performance characteristics are generally worse than those of the non-augmented dataset.
However, an important point can be observed: the performance after the pre-training (particularly for the validation set) does not differ to the same extent as it did with non-augmented data.
As with the non-augmented dataset, the baseline loss (the rightmost point) exhibits the highest validation set loss.
There, the difference between the lowest and the highest subset losses has been tenfold, while it is roughly the same for all subset fractions with the augmented data.

The ten times larger size of the augmented dataset leads to overdetermination ratios $Q$ mostly (except for the fraction $\nicefrac{1}{128}$) over unity.
Then, even the small-fraction subsets generalize acceptably (which is the goal of sufficient overdetermination).

\subsection{\uppercase{Dataset \cifar}}\label{sec:results_cifar}

\begin{figure}
	\centering
	\includegraphics[height=5cm]{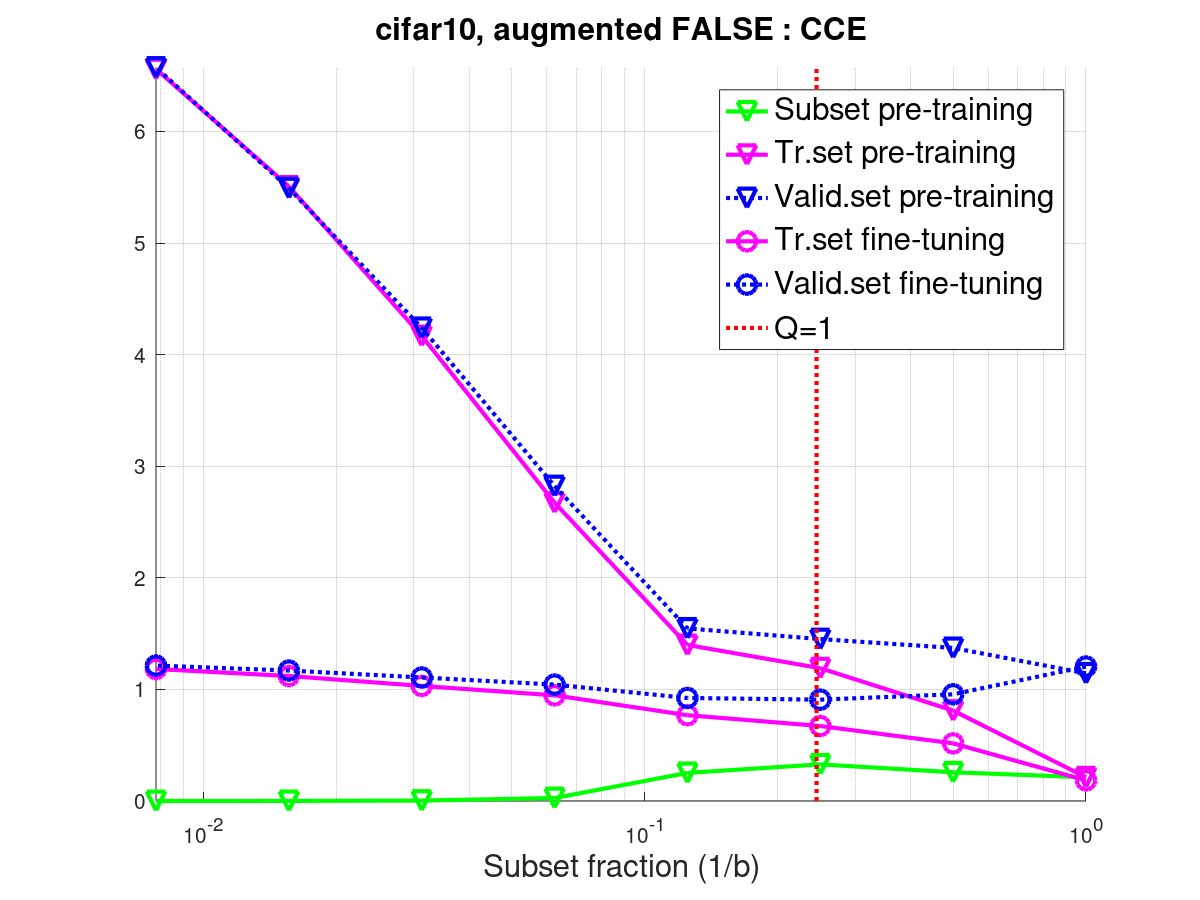}
	\caption{Dataset \cifar{} (non-augmented), loss optima for a pre-trained subset and the whole training set in dependence from the subset size (as a fraction of the training set).}\label{fig:Loss_cifar10_naug}
\end{figure}
\begin{figure}
	\centering
	\includegraphics[height=5cm]{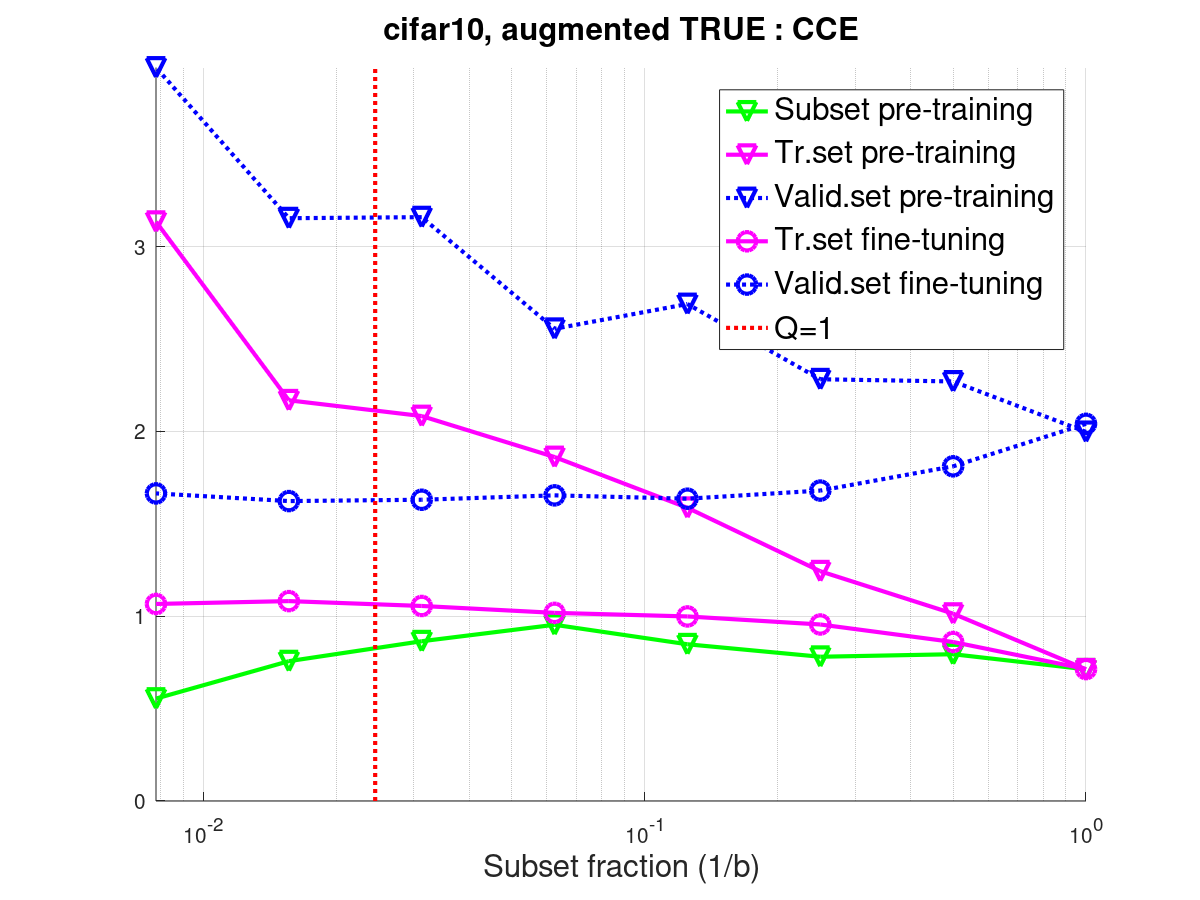}
	\caption{Dataset \cifar{} (augmented), loss optima for a pre-trained subset and the whole training set dependent on the subset size (as a fraction of the training set).}\label{fig:Loss_cifar10_aug}
\end{figure}

The results for non-augmented \cifar{}  data are depicted in~\cref{fig:Loss_cifar10_naug}, those for augmented data in~\cref{fig:Loss_cifar10_aug}.
Due to the size of \cifar{} being close to \mnist{}, the overdetermination ratios are also very similar.
Since \cifar{} is substantially harder to classify, losses and accuracies are worse.

The accuracy is a secondary characteristic (since the categorical cross-entropy is minimized), and its explanatory power is limited.
For this and the space reasons, accuracies will not be presented for \cifar{} and \cifarb{}.

Nevertheless, the conclusions are similar to those from \mnist{}.
The gap between the subset's and the entire training set's fine-tuning performance diminishes as the subset grows.
This gap is large with non-augmented data in~\cref{fig:Loss_cifar10_naug} because of low to overdetermination ratios $Q$ but substantially smaller for augmented data in~\cref{fig:Loss_cifar10_aug} where overdetermination ratios are sufficient.
The verification set performance after both training phases is typically better with subsets of most sizes than with the whole training set.

\subsection{\uppercase{Dataset \cifarb{}}}\label{sec:results_cifar100}
\begin{figure}
	\centering
	\includegraphics[height=5cm]{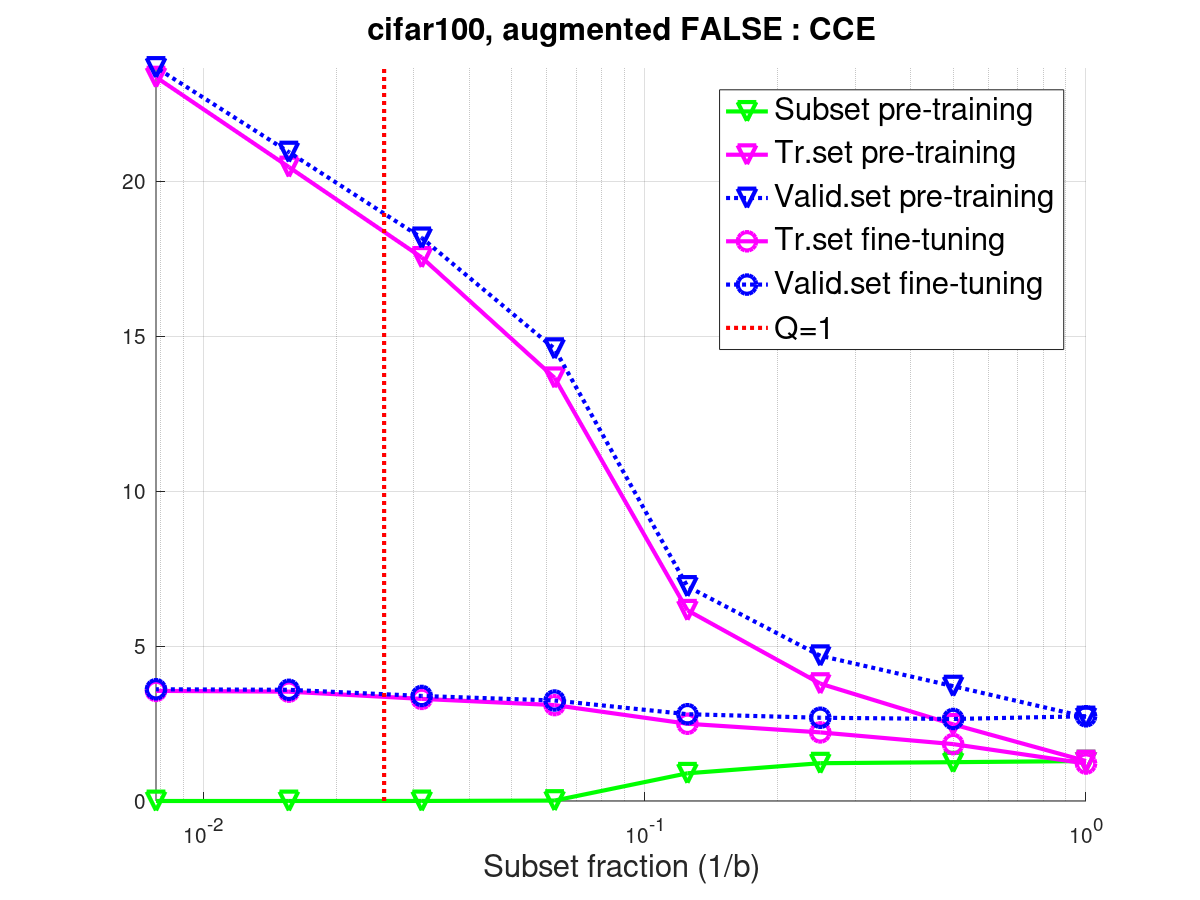}
	\caption{Dataset \cifarb{} (non-augmented), loss optima for a pre-trained subset and the whole training set in dependence from the subset size (as a fraction of the training set).}\label{fig:Loss_cifar100_naug}
\end{figure}

The results for (non-augmented) \cifarb{} data are depicted in~\cref{fig:Loss_cifar100_naug}. 
This classification task differentiates \num{100} classes so that there are only \num{500} examples per class.
Optimum losses for this benchmark are higher than for the previous ones.

For small subset fractions, the representation of the classes is probably insufficient.
This may explain the large gap between the subset loss and the training set loss after the subset training with small subset fractions.
These may contain, on average, even as few examples per class as four.
Nevertheless, the loss for the validation set with various subset sizes is close to the baseline loss for the conventional full-size training.

\section{\uppercase{Conclusion}}\label{sec:discussion}
\begin{figure}
	\centering
	\includegraphics[height=5cm]{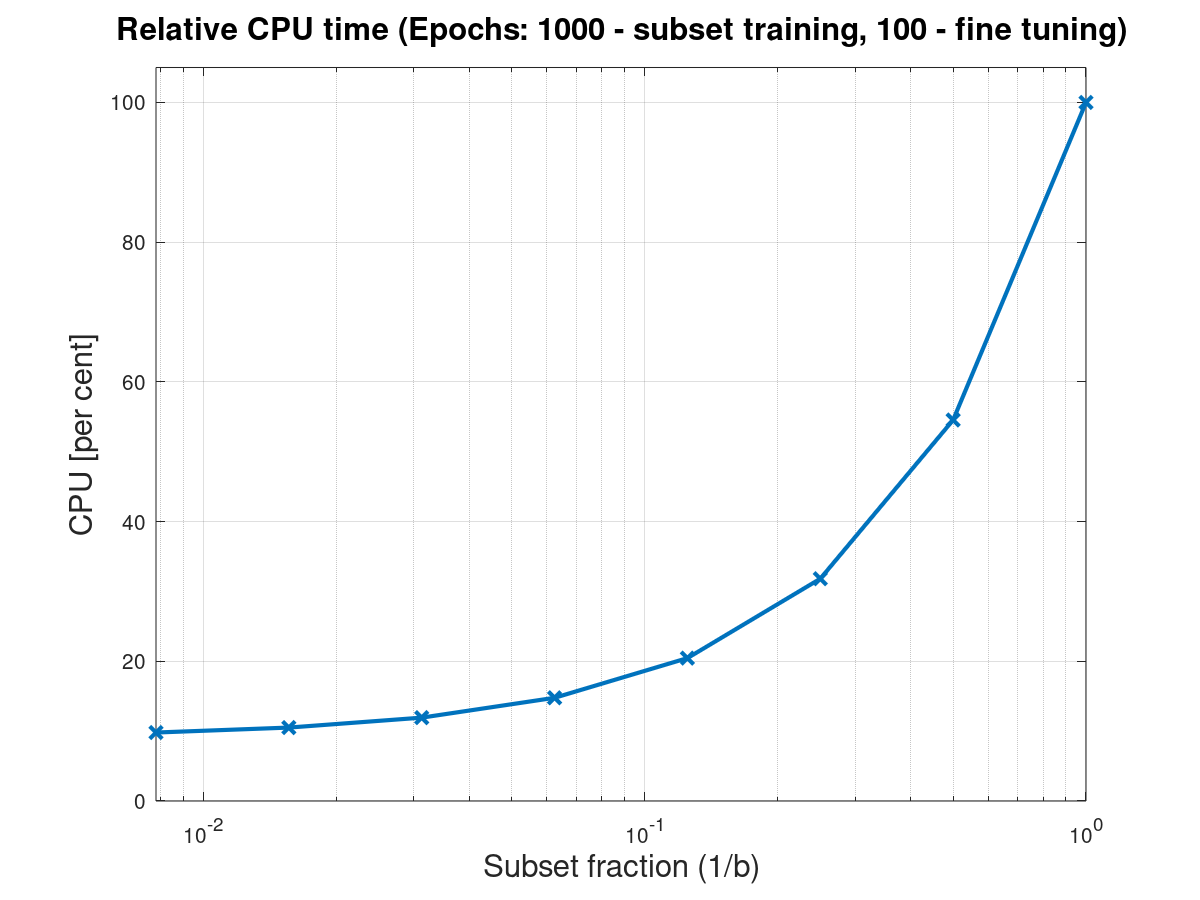}
	\caption{Training time relative to the conventional training in dependence from the subset size (in percent).}\label{fig:relative_cpu_time}
\end{figure}

The experiments presented support the concept of subset training.
We demonstrated the following elements.
\begin{itemize}
	\item
	      The subset training leads to results comparable with conventional training over the whole training set.
	\item
	      The overdetermination ratio $Q$ (preferably above unity) should determine the subset size.
	      Nevertheless, even underdetermined subsets may lead to a good fine-tuning result, although they put more workload on the fine-tuning (to close the large generalization gap).
	\item
	      To summarize, even small subsets can be representative enough to approximate the training set loss minimum well whenever the overdetermination ratio sufficiently exceeds unity.
\end{itemize}
The most important achievement is the reduction of computing expenses.
Most optimizing iterations are done on the subset, where the computational time per epoch is a fraction of that for the whole training set.
In our experiments with ten times more subset training epochs than fine-tuning epochs, the relative computing time in percent of the baseline is shown in~\cref{fig:relative_cpu_time}.
Computational resource savings of \SI{90}{\percent} and more are possible.

This empirical evaluation using five benchmarks from the CV domain is insufficient for making general conclusions.
Large datasets such as ImageNet are to be tested in the future.
They have been omitted because many experiments are necessary to produce sufficient statistics.
Furthermore, these experiments can be extended to language benchmarks and language models.

It is also important to investigate the behavior of the second-order optimization algorithms such as conjugate gradient~\cite{hestenes_methods_1952,fletcher_function_1964}.
Their strength can develop only with a sufficient number of iterations.
This is an obstacle if very large training sets are a part of the task.
Appropriately chosen subsets can make such training feasible and help to reach good performance even with models of moderate size.

\bibliographystyle{apalike}
{\small \bibliography{references}}

\end{document}